\documentclass[letterpaper]{article} 
\usepackage{aaai24}  
\usepackage{times}  
\usepackage{helvet}  
\usepackage{courier}  
\usepackage[hyphens]{url}  
\usepackage{graphicx} 
\usepackage{natbib}  
\usepackage{caption} 
\usepackage{algorithm}
\usepackage{algorithmic}
\usepackage{multirow}
\usepackage{adjustbox}
\usepackage{amssymb}
\usepackage{newfloat}
\usepackage{listings}
\DeclareCaptionStyle{ruled}{labelfont=normalfont,labelsep=colon,strut=off} 
\floatstyle{ruled}
\newfloat{listing}{tb}{lst}{}
\floatname{listing}{Listing}
\title{U-Mixer: An Unet-Mixer Architecture with Stationarity Correction for Time Series Forecasting}
\author{
    Xiang Ma\textsuperscript{\rm 1},
    Xuemei Li\textsuperscript{\rm 1},
    Lexin Fang\textsuperscript{\rm 1},
    Tianlong Zhao\textsuperscript{\rm 1},
    Caiming Zhang\textsuperscript{\rm 1,\rm 2}\thanks{Corresponding author.}
}
\affiliations{
    \textsuperscript{\rm 1}School of Software, Shandong University, Jinan 250101, China\\
    \textsuperscript{\rm 2}Shandong Provincial Laboratory of Future Intelligence and Financial Engineering, Yantai 264005, China\\

   \{xiangma, fanglexin, tianlongzhao\}@mail.sdu.edu.cn, \{xmli, czhang\}@sdu.edu.cn
}

\usepackage{bibentry}

\begin{document}

\maketitle
\begin{abstract}
	Time series forecasting is a crucial task in various domains. Caused by factors such as trends, seasonality, or irregular fluctuations, time series often exhibits non-stationary. It obstructs stable feature propagation through deep layers, disrupts feature distributions, and complicates learning data distribution changes. As a result, many existing models struggle to capture the underlying patterns, leading to degraded forecasting performance. In this study, we tackle the challenge of non-stationarity in time series forecasting with our proposed framework called U-Mixer. By combining Unet and Mixer, U-Mixer effectively captures local temporal dependencies between different patches and channels separately to avoid the influence of distribution variations among channels, and merge low- and high-levels features to obtain comprehensive data representations. The key contribution is a novel stationarity correction method, explicitly restoring data distribution by constraining the difference in stationarity between the data before and after model processing to restore the non-stationarity information, while ensuring the temporal dependencies are preserved. Through extensive experiments on various real-world time series datasets, U-Mixer demonstrates its effectiveness and robustness, and achieves 14.5\% and 7.7\% improvements over state-of-the-art (SOTA) methods.
\end{abstract}

\section{Introduction}
Time series forecasting plays a vital role in various domains such as finance \cite{mine_finance}, weather forecasting \cite{ltt_weather}, and sensor data analysis \cite{ztl3_sensor}. Extracting meaningful patterns, understanding the underlying dynamics of time series to forecast future trends are crucial for informed decision-making and effective problem-solving \cite{zf_1}. With the advent of deep learning, convolutional neural networks (CNNs) \cite{cnn} and Transformers \cite{transformer} have shown remarkable progress in capturing temporal dependencies and extracting features from time series. Recently, the Mixer architecture \cite{mixer}, initially introduced for vision tasks, has gained attention for its ability to model complex relationships within sequential data. However, the application of Mixer to time series forecasting also presents challenges. Time series data often exhibits non-stationary caused by factors such as trend, seasonality, or irregular fluctuations. Such property can hinder accurate modeling and prediction, thereby limiting the effectiveness of the Mixer architecture. In dealing with non-stationary, Mixer mainly have the following three problems: (1) The deep structure of Mixer leads to unstable propagation of shallow features. When information flows through multiple layers, the transmission of low-level features becomes non-stationary. (2) The mixed feature extraction of different channels leads to non-stationary feature distributions, due to significant distribution variations among channels. (3) Model training cannot intuitively learn changes in data distribution, resulting in shifted and non-stationary distribution of predicted values.

In this article, we present U-Mixer to address the issue of non-stationarity in time series forecasting. U-Mixer combines the advantages of both Unet and Mixer architecture to capture local temporal patterns of different levels while separately handles the temporal and channel interactions. The key contribution of U-Mixer lies in the novel stationarity correction technique, which explicitly restores the distribution of data by correcting the stationarity of data.

Specifically, we divide the time series into some patches and process them independently using the Mixer architecture. This patch-based processing allows localized analysis of temporal patterns and captures fine-grained details within the data. Mixer effectively captures the dependencies between different patches and channels without disrupting the overall temporal sequence through the multi-layer perceptron (MLP) block, and learns meaningful representations from the entire time series. Simultaneously, it separately handles the temporal interactions and channel interactions, avoiding feature instability caused by significant variations in data distributions across different channels. Different from MLP-Mixer, U-Mixer adopts a Unet architecture to merge the low- and high-level features and obtain a more comprehensive and richer data representation, thereby improving the ability to understand and model data. Additionally, in order to improve model performance, time series often undergoes a process of stationarization. However, due to the loss of non-stationary information, the model cannot intuitively learn the changes in data distribution, resulting in a shift in the distribution of predicted values. Learning the mapping of data distribution through training is inherently a challenge. Therefore, we propose a stationarity correction method by constraining the difference in stationarity between the data before and after model processing to restore the non-stationary information, while consider the temporal dependencies of the data. 

Our contributions are summarized as
\begin{itemize}
	\item We propose U-Mixer a new time series forecasting framework that can capture local temporal patterns of different levels and handle the temporal and channel interactions separately.
	\item We propose a stationarity correction method that explicitly restores the non-stationarity information of data by constraining the stationarity differences before and after model processing, while ensuring temporal dependencies within the data.
	\item U-Mixer adopts the Unet architecture to merge the low- and high-level features, which enables more comprehensive and richer representations of the data.
	\item We demonstrate the effectiveness and robustness of U-Mixer through extensive experiments on various real-world time series datasets.    
\end{itemize}

\section{Related Work}
\subsection{Time Series Prediction}
Time series forecasting is the task to predict future values for one or more variables based on a set of historical observations. Traditional statistical methods and machine learning methods have been widely used but struggle with complex nonlinear patterns, although they are simple and interpretable. In recent years, deep learning models have been widely investigated for this task. Especially long short-term memory (LSTM) \cite{lstm}, a well-known variant of recurrent neural networks (RNNs), is widely employed in time series forecasting. However, they may suffer from vanishing or exploding gradients, limiting their ability to capture long-term dependencies. Convolutional Neural Networks (CNNs) \cite{ztl3_sensor} have also been adapted for this task, by extracting local patterns and identify relevant features across different time steps. CNN-based models have shown promising results in capturing spatial and temporal dependencies within the data.

More recently, Transformer-based models \cite{Informer,Autoformer,Fedformer} have gained attention in time series forecasting. Transformers leverage self-attention mechanisms to effectively capture global dependencies and long-range interactions within the data. They have achieved state-of-the-art (SOTA) results in various time series forecasting tasks. In contrast to expectations, recent studies \cite{dlinear} have revealed that even a basic univariate linear model can outperform complex multivariate Transformer models by a margin on widely-used long-term prediction benchmarks. Despite the significant advancements in Transformer-based models, this unexpected finding highlights the effectiveness of linear models. In this paper, we introduce the Mixer architecture for time series forecasting to fully utilize the performance of linear models, which is designed by stacking MLPs. And we combine Mixer with the Unet architecture to integrate different levels of features to build more comprehensive richer representations
\subsection{Stationarization for Time Series Forecasting}
Stationarization serves as a fundamental assumption for time series analysis, allowing us to apply various models to efficiently capture patterns in the data and enhence the robustness of models. To stabilize time series data, traditional methods employ various preprocessing approaches, such as differencing, to remove trends, seasonality, and non-stationarity in the data. As for deep models, the presence of non-stationarity and the accompanying variation in data distributions pose significant challenges to time series forecasting. To address this, stationarization are commonly explored and employed as a preprocessing step for deep model inputs \cite{Non-stationary}. By transforming the data into a more stationary form, these methods aim to mitigate the difficulties associated with non-stationarity and enable more effective training and prediction with deep models. Normalization is a widely adopted stationarization method, aiming to mitigate the negative effects of non-stationary features on the learning process by transforming the data into a range that is better aligned with the model. 

While applying normalization can address the issue of non-stationarity, it introduces a potential preblem \cite{revin}. Normalization can inadvertently remove non-stationary information that may hold valuable insights for predicting future values. Because normalization changes the distribution of features, potentially hindering the model`s ability to capture the nuanced dynamics of the time series. Some research \cite{revin,Non-stationary} explicitly return the information deleted through input normalization to the model, eliminating the need for the model to reconstruct the original distribution, thereby reducing the difficulty of modeling. However, directly stationarizing time series will damage the model’s capability of modeling specific temporal dependency. Therefore, we propose a stationarity correction method by constraining the difference in stationarity between the data before and after model processing. This method explicitly conveys statistical features of data distribution while maintaining temporal dependencies whthin the data.
\begin{figure}[t]
	\centering
	\includegraphics[width=0.4\textwidth]{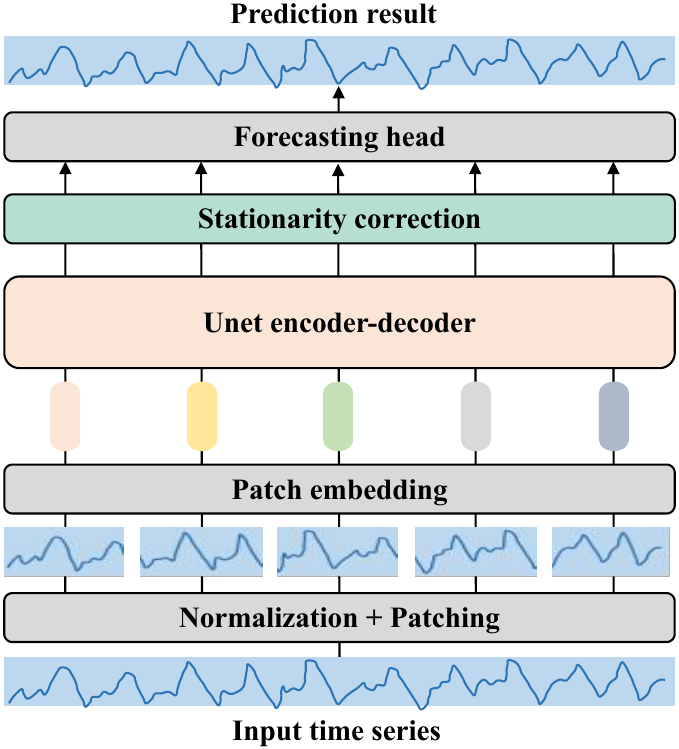} 
	\caption{The architecture of U-Mixer, which consists of per-patch embeddings, Unet encoder-decoder, stationarity correction and a forecasting head.}
	\label{framework}
\end{figure}

\section{Methodology}
The architecture of U-Mixer is shown in Figure \ref{framework}, and the details are described in the following sections.
\subsection{Notations}
Define the historical data as $X \in \mathbb{R}^{C\times L}=\{x_i\ |\ i\in[1,L]\}$. Here $L$ means the length of the input sequence and $C$ is the number of channels (or variables). $x_i$ is a vector of dimension $C$ at time step $i$. Let the ground truth future values be $Y \in \mathbb{R}^{C\times H}=\{x_i\ |\ i\in[L+1,L+H]\}$. Here $H$ means the length of forecasting sequence. We focus on using a learning model $\mathcal{M}$ to analysis $X$ and achieve predicting the values of $Y$, and the process can be expressed as $\hat{Y} = \mathcal{M}(X)$. $\hat{Y}$ means the prediction results. 
\subsection{Normalization and Patch Embedding}
The input data $X$ go through a normalization process, which is: $X=\frac{X-\mu_{in}}{\sigma_{in}}$. Here $\mu_{in}$ and $\sigma_{in}$ is the vectors composed of the mean and variance of all channels in $X$, respectively. Through this process, the range of data variation is adjusted to a more suitable scale, contributing to enhanced stability and performance of the model. 

After normalization, $X$ is divided into overlapping or non-overlapping patches. Define the patch length as $P$ and the stride between two consecutive patches as $S$. We can obtain the patch sequence $X_p\in \mathbb{R}^{(C\times N)\times P}$. Here $N=\lfloor\frac{L-P}{S} \rfloor+2$ is the number of patches. $\lfloor\cdot \rfloor$ is floor function. We repeat the last column of $X$ $S$ times and pad it to the original sequence before patching.

Patches are mapped to the embeddings $X_d\in \mathbb{R}^{(C\times N)\times D}$, where $D$ is the latent space dimension. We use a linear projection $W_{val}\in \mathbb{R}^{P\times D}$ to learn the mapping relationship and a additive position encoding $W_{pos}\in \mathbb{R}^{(C\times N)\times D}$ to provide the information about the relative position of patches. Thus, $X_d=X_pW_{val}+W_{pos}$. Then $X_d$ will be feeded into the Unet encoder-decoder to capture the dependencies between different patches and channels. Because model processing will cause the distribution in the patches to change, here we also need to record the mean $\mu_{x}$, the variance $\sigma_{x}$ and autocorrelation matrix $R(X_d)$ of $X_d$, so that the distribution of the model`s output will be restored in the subsequent stationarity correction operation.
\begin{figure}[t]
	\centering
	\includegraphics[width=0.43\textwidth]{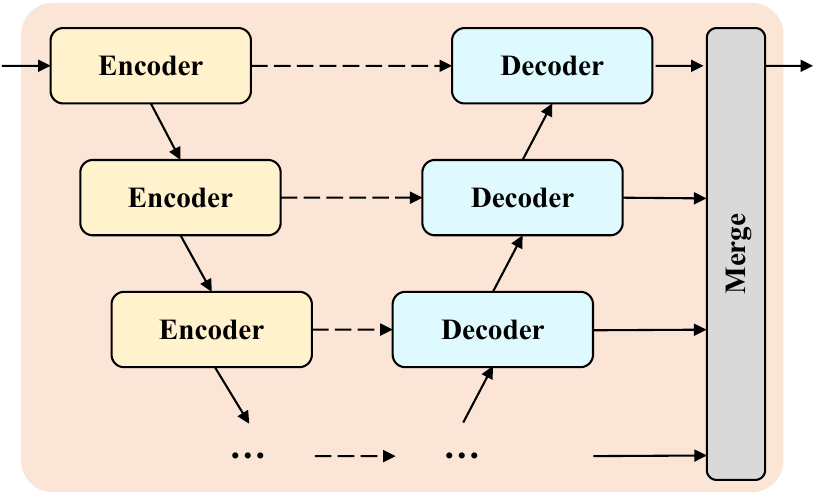} 
	\caption{The Unet encoder-decoder of U-Mixer. Encoders and decoders are both MLP blocks. The term "Merge" refers to the combining process of features from different levels.}
	\label{mmixerl}
\end{figure}

\subsection{Unet Encoder-decoder}
As shown in Figure \ref{mmixerl}, U-Mixer introduces a novel time series prediction network that combines the Unet architecture with the Mixer architecture. 

U-Mixer adopts the encoder-decoder structure of Unet with multiple levels. Encoders adopt a hierarchical structure, progressively extracting features of low- and high-levels from embeddings $X_d$. Each encoder is responsible for transforming the input embeddings into high-dimensional representations that captures key features and contextual information. Define the input of encoders as:
\begin{equation}
	X_{in,i}=\left\{
	\begin{array}{lll}
		X_d &{i=1}\\
		X_{out,i-1}&{i\in (1,M]}
	\end{array} \right.
\end{equation}
$X_{in,i}$ means the input of the i-level encoder, and $M$ is the number of levels. The output of encoders can be expressed as:
\begin{equation}
	X_{out,i}=\mathcal{M}_{en,i}(X_{in,i}), \ \ i\in [1,M]
\end{equation}
$X_{out,i}$ is the output of the i-level encoder and $\mathcal{M}_{en,i}$ refers to the i-level encoder.

Decoders also adopt a hierarchical structure, progressively analyzing representations generated by encoders. Each decoder is responsible for generating the parsed representations by analyzing the representations from the output of the previous decoder. During the parsing process, each decoder also need to consider the output of the same level encoder to preserve and utilize the features of the same level. This process is achieved through the skip connections at the corresponding level. The input of decoders can be formalized as:
\begin{equation}
	Y_{in,i}=\left\{
	\begin{array}{lll}
		Y_{out,i+1}&{i\in [1,M)}\\
		X_{out,i} &{i=M}
	\end{array} \right.
\end{equation}
$Y_{in,i}$ means the input of the i-level decoder. The output of decoders is defined as:
\begin{small} 
\begin{equation}
	Y_{out,i}=\\
	\left\{
	\begin{array}{ll}
	W_y(\mathcal{M}_{de,i}(Y_{out,i+1})+\mathcal{M}_{de,i}(Y_{in,i}))&{i\in [1,M)}\\
	\mathcal{M}_{de,i}(X_{out,M}) &{i=M}
	\end{array} \right.
\end{equation}
\end{small}
$Y_{out,i}$ is the output of the i-level decoder and $\mathcal{M}_{de,i}$ refers to the i-level decoder. $W_y$ is a simple linear layer to merge the features generated from the same level encoder and the previous decoder. 

For ease of description, the final output of the encoder-decoder structure is defined as $Y_d\in \mathbb{R}^{(C\times N)\times D}=Y_{out,1}$.
\begin{figure*}[t]
	\centering
	\includegraphics[width=0.98\textwidth]{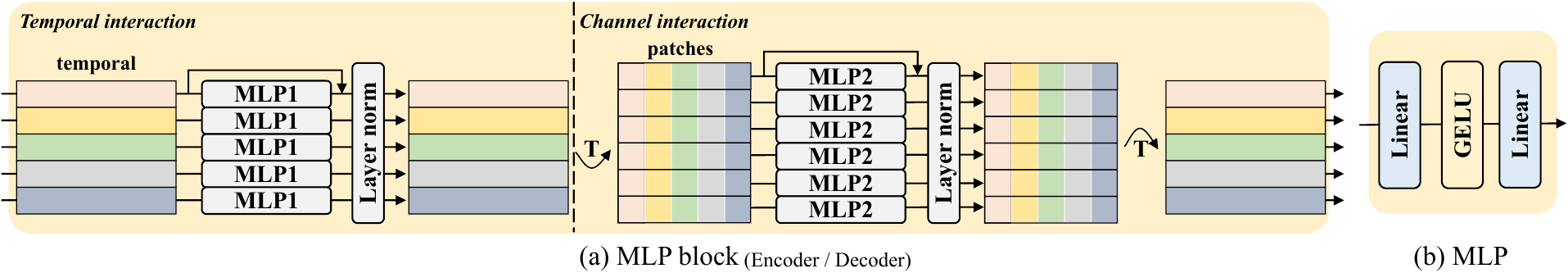} 
	\caption{The MLP block. (a) is the MLP block, which contains one temporal MLP layer and one channel MLP layer. (b) is the specific structure of MLP, which consists of two fully-connected layers and a GELU nonlinearity.}
	\label{mlpblock}
\end{figure*}

\subsection{MLP Block}
Encoders and decoders are both MLP blocks, as shown in Figure \ref{mlpblock}(a). MLP block contains two MLP layers, which are used to implement temporary interaction and channel interaction, respectively. One MLP layer is first used to perform interactions on the input data in the temporal dimension, which contains several MLPs for independently processing each channel. This process has no interactions between different channels to avoid affecting temporal interactions due to differences in distribution between channels. After layer normalization, the output is transposed. Another MLP layer is used to perform interaction on the channel dimension. The output is transposed back to its previous shape after layer normalization. During the interaction process, we also employ skip connections to concatenate the input and output features of MLP layers, reducing the information loss introduced by feature transformation and providing a more comprehensive feature representation. An MLP consists of two linear layers, a GELU activation function and a dropout, as shown in Figure \ref{mlpblock}(b). As the core components of MLP, linear layers perform linear transformations on the inputs to learn linear combinations of features. GELU achieves non-linear mapping by applying a Gaussian error linear transformation to the inputs, which is widely used in time series forecasting. Dropout improves the model`s robustness by reducing the sensitivity of the network to specific features or input patterns. 
\subsection{Stationarity Correction}
Due to the removal of non-stationary information from the input, the distribution of the data goes through a significant shift. Existing methods for restoring the data distribution primarily focus on statistical measures such as mean and variance, without considering the temporal dependencies within the data. As a result, essential features like trends and seasonal in the original data may be affected. Therefore, we introduce a novel stationarity correction method that constraining the relationship between the stationarity of the time series before and after model processing to rectify the data distribution while preserving data dependencies. 

The stationarity of time series is mainly characterized by two aspects: the mean and the covariance. The mean is mainly responsible for constraining the distribution from the perspective of statistics, while covariance constrains the distribution from the perspective of temporal dependencies. The change in the mean of a data distribution is typically achieved through global addition or subtraction operations and does not affect the covariance. However, adjusting the covariance can lead to changes in the data distribution's mean. Therefore, we first adjust the covariance of the data. The covariance $Cov(X_d,X_d^{i})$ can represent the dependence between the series $X_d$ and its i lag series, which is not sufficient to fully describe the temporal dependencies of the entire time series. Therefore, we introduce autocorrelation matrice to provide a more comprehensive constraint on the time series dependencies.

Define $R(X_d)\in \mathbb{R}^{L\times L}=\{R_{i,j}(X_d)\ |\ i,j\in[1,L]\}$ as the autocorrelation matrice of $X_d$, and $R_{i,j}(X_d)=\frac{Cov(X_d^{i},X_d^{j})}{\sqrt{\sigma_x^{i}\sigma_x^{j}}}$. Here $Cov(\cdot)$ is the computation of covariance. $X_d^{i}$ means the $i$ lag series of $X_d$, and $\sigma_d^{i}$ represents the variance of $X_d^{i}$. We perform an affine transformation on the output of model $Y_d$ using matrix $\alpha$, such that the autocorrelation matrix $R(\alpha Y_d)$ approximates $R(X_d)$. Here $\alpha=[\alpha_1,\alpha_2,\cdots,\alpha_L]\in\mathbb{R}^{1\times L}$ is a matrix, and $\alpha_i$ is a scalar. The constraint between $R(X_d)$ and $R(\alpha Y_d)$ can be expressed as:
\begin{equation}
	\mathcal{M}_{\alpha}=\Vert R(X_d)-R(\alpha Y_d) \Vert_F^2
\end{equation}
$Y_d$ is a non-stationary time series, multiplying $Y_d$ by $\alpha$ will result in the changes between data points in $Y_d$ being scaled by a factor of $\alpha^2$. So $\mathcal{M}_{\alpha}$ is equivalent to:
\begin{equation}
	\mathcal{M}_{\alpha}=\Vert R(X_d)-\alpha^2R(Y_d) \Vert_F^2
\end{equation}
Here:
\begin{equation}
	\alpha_i=\sqrt{\frac{\sum_{j=1}^{L}R_{i,j}(X_d)R_{i,j}(Y_d)}{\sum_{j=1}^{L}R_{i,j}^2(X_d)}} 
\end{equation}
According to the Wiener-Khinchin theorem \cite{Wiener-Khinchin}, we can accelerate the computation of $\alpha_i$ by the Fast Fourier Transform (FFT):
\begin{equation}
	 \alpha_i=\sqrt{\frac{\sum_{j=1}^{L}\mathcal{F}^{-1}(\mathcal{F}(\overline{X_d^i})\overline{\mathcal{F}(\overline{X_d^j}}))\mathcal{F}^{-1}(\mathcal{F}(\overline{Y_d^i})\overline{\mathcal{F}(\overline{Y_d^j}}))}{\sum_{j=1}^{L}\mathcal{F}^{-1}(\mathcal{F}(\overline{X_d^i})\overline{\mathcal{F}(\overline{X_d^j}}))^2}}
\end{equation}
The means of $X_d$ and $Y_d$ need to be zero to satisfy the Wiener-Khinchin theorem. Here $\overline{X_d^i}=X_d^i-\mu_x^i$ and $\overline{Y_d^i}=Y_d^i-\mu_y^i$. $\mu_x^i$ and $\mu_y^i$ are the means of $X_d^i$ and $Y_d^i$, respectively. $\alpha$ is determined by $\overline{X_d}$ and $\overline{Y_d}$. 

The output $Y_d$ can be update to $\hat{Y_d}\in \mathbb{R}^{(C\times N)\times D}=\alpha Y_d+\bigtriangledown_\mu$, $\bigtriangledown_\mu=\mu_x-\mu_y$ is used to adjust the difference in mean between $Y_d$ and $X_d$.
\begin{table*}[th]
	\renewcommand{\arraystretch}{0.8}
	\centering
	\resizebox{\textwidth}{!}{\setlength{\tabcolsep}{0.38mm}{
			\begin{tabular}{c c c c|c c|c c|c c|c c|c c|c c|c c}
				\hline
				\multicolumn{2}{c}{Models}&\multicolumn{2}{|c}{\textbf{U-Mixer}}&\multicolumn{2}{c}{TimesNet}&\multicolumn{2}{c}{ETSformer}&\multicolumn{2}{c}{LightTS}&\multicolumn{2}{c}{DLinear}&\multicolumn{2}{c}{FEDformer}&\multicolumn{2}{c}{Autoformer}&\multicolumn{2}{c}{LSSL}\\
				\cline{1-18}
				\multicolumn{2}{c|}{Metric}&MSE&MAE&MSE&MAE&MSE&MAE&MSE&MAE&MSE&MAE&MSE&MAE&MSE&MAE&MSE&MAE\\
				\hline
				\multicolumn{1}{c|}{\multirow{5}{*}{\rotatebox{90}{ETTm1}}}&\multicolumn{1}{c|}{96}&\textbf{0.317$\pm$2e$^{-3}$}&\textbf{0.349$\pm$2e$^{-3}$}&0.338&0.375&0.375&0.398&0.374&0.400&0.345&0.372&0.379&0.419&0.505&0.475&0.450&0.477\\
				\multicolumn{1}{c|}{}&\multicolumn{1}{c|}{192}&\textbf{0.369$\pm$2e$^{-3}$}&\textbf{0.376$\pm$2e$^{-3}$}&0.374&0.387&0.408&0.410&0.400&0.407&0.380&0.389&0.426&0.441&0.553&0.496&0.469&0.481\\
				\multicolumn{1}{c|}{}&\multicolumn{1}{c|}{336}&\textbf{0.395$\pm$3e$^{-3}$}&\textbf{0.393$\pm$4e$^{-3}$}&0.410&0.411&0.435&0.428&0.438&0.438&0.413&0.413&0.445&0.459&0.621&0.537&0.583&0.574\\
				\multicolumn{1}{c|}{}&\multicolumn{1}{c|}{720}&\textbf{0.443$\pm$2e$^{-3}$}&\textbf{0.424$\pm$1e$^{-3}$}&0.478&0.450&0.499&0.462&0.527&0.502&0.474&0.453&0.543&0.490&0.671&0.561&0.632&0.596\\
				\cline{2-18}
				\multicolumn{1}{c|}{}&\multicolumn{1}{c|}{Avg}&\textbf{0.381$\pm$2e$^{-3}$}&\textbf{0.386$\pm$3e$^{-3}$}&0.400&0.406&0.429&0.425&0.435&0.437&0.403&0.407&0.448&0.452&0.588&0.517&0.533&0.532\\
				\hline
				\multicolumn{1}{c|}{\multirow{5}{*}{\rotatebox{90}{ETTm2}}}&\multicolumn{1}{c|}{96}&\textbf{0.178$\pm$2e$^{-3}$}&\textbf{0.256$\pm$2e$^{-3}$}&0.187&0.267&0.189&0.280&0.209&0.308&0.193&0.292&0.203&0.287&0.255&0.339&0.243&0.342\\
				\multicolumn{1}{c|}{}&\multicolumn{1}{c|}{192}&\textbf{0.243$\pm$3e$^{-3}$}&\textbf{0.301$\pm$2e$^{-3}$}&0.249&0.309&0.253&0.319&0.311&0.382&0.284&0.362&0.269&0.328&0.281&0.340&0.392&0.448\\
				\multicolumn{1}{c|}{}&\multicolumn{1}{c|}{336}&0.331$\pm$1e$^{-3}$&0.355$\pm$2e$^{-3}$&0.321&\textbf{0.351}&\textbf{0.314}&0.357&0.442&0.466&0.369&0.427&0.325&0.366&0.339&0.372&0.932&0.724\\
				\multicolumn{1}{c|}{}&\multicolumn{1}{c|}{720}&0.434$\pm$3e$^{-3}$&0.413$\pm$2e$^{-3}$&\textbf{0.408}&\textbf{0.403}&0.414&0.413&0.675&0.587&0.554&0.522&0.421&0.415&0.433&0.432&1.372&0.879\\
				\cline{2-18}
				\multicolumn{1}{c|}{}&\multicolumn{1}{c|}{Avg}&0.291$\pm$3e$^{-3}$&\textbf{0.331$\pm$2e$^{-3}$}&\textbf{0.291}&0.333&0.293&0.342&0.409&0.436&0.350&0.401&0.305&0.349&0.327&0.371&0.735&0.598\\
				\hline
				\multicolumn{1}{c|}{\multirow{5}{*}{\rotatebox{90}{ETTh1}}}&\multicolumn{1}{c|}{96}&\textbf{0.370$\pm$9e$^{-4}$}&\textbf{0.390$\pm$1e$^{-3}$}&0.384&0.402&0.494&0.479&0.424&0.432&0.386&0.400&0.376&0.419&0.449&0.459&0.548&0.528\\
				\multicolumn{1}{c|}{}&\multicolumn{1}{c|}{192}&0.423$\pm$1e$^{-3}$&\textbf{0.421$\pm$1e$^{-3}$}&0.436&0.429&0.538&0.504&0.475&0.462&0.437&0.432&\textbf{0.420}&0.448&0.500&0.482&0.542&0.526\\
				\multicolumn{1}{c|}{}&\multicolumn{1}{c|}{336}&0.470$\pm$2e$^{-3}$&\textbf{0.442$\pm$2e$^{-3}$}&0.491&0.469&0.574&0.521&0.518&0.488&0.481&0.459&\textbf{0.459}&0.465&0.521&0.496&1.298&0.942\\
				\multicolumn{1}{c|}{}&\multicolumn{1}{c|}{720}&\textbf{0.500$\pm$3e$^{-3}$}&\textbf{0.473$\pm$4e$^{-3}$}&0.521&0.500&0.562&0.535&0.547&0.533&0.519&0.516&0.506&0.507&0.514&0.512&0.721&0.659\\
				\cline{2-18}
				\multicolumn{1}{c|}{}&\multicolumn{1}{c|}{Avg}&\textbf{0.440$\pm$3e$^{-3}$}&\textbf{0.432$\pm$3e$^{-3}$}&0.458&0.450&0.542&0.510&0.491&0.479&0.456&0.452&0.440&0.460&0.496&0.487&0.777&0.664\\
				\hline
				\multicolumn{1}{c|}{\multirow{5}{*}{\rotatebox{90}{ETTh2}}}&\multicolumn{1}{c|}{96}&\textbf{0.290$\pm$4e$^{-3}$}&\textbf{0.335$\pm$3e$^{-3}$}&0.340&0.374&0.340&0.391&0.397&0.437&0.333&0.387&0.358&0.397&0.346&0.388&1.616&1.036\\
				\multicolumn{1}{c|}{}&\multicolumn{1}{c|}{192}&\textbf{0.366$\pm$4e$^{-3}$}&\textbf{0.386$\pm$2e$^{-3}$}&0.402&0.414&0.430&0.439&0.520&0.504&0.477&0.476&0.429&0.439&0.456&0.452&2.083&1.197\\
				\multicolumn{1}{c|}{}&\multicolumn{1}{c|}{336}&\textbf{0.423$\pm$2e$^{-3}$}&\textbf{0.428$\pm$2e$^{-3}$}&0.452&0.452&0.485&0.479&0.626&0.559&0.594&0.541&0.496&0.487&0.482&0.486&2.970&1.439\\
				\multicolumn{1}{c|}{}&\multicolumn{1}{c|}{720}&\textbf{0.446$\pm$2e$^{-3}$}&\textbf{0.445$\pm$3e$^{-3}$}&0.462&0.468&0.500&0.497&0.863&0.672&0.831&0.657&0.463&0.474&0.515&0.511&2.576&1.363\\
				\cline{2-18}
				\multicolumn{1}{c|}{}&\multicolumn{1}{c|}{Avg}&\textbf{0.381$\pm$4e$^{-3}$}&\textbf{0.399$\pm$3e$^{-3}$}&0.414&0.427&0.439&0.452&0.602&0.543&0.559&0.515&0.437&0.449&0.450&0.459&2.311&1.259\\
				\hline
				\multicolumn{1}{c|}{\multirow{5}{*}{\rotatebox{90}{Electricity}}}&\multicolumn{1}{c|}{96}&\textbf{0.151$\pm$2e$^{-4}$}&\textbf{0.240$\pm$1e$^{-3}$}&0.168&0.272&0.187&0.304&0.207&0.307&0.197&0.282&0.193&0.308&0.201&0.317&0.300&0.392\\
				\multicolumn{1}{c|}{}&\multicolumn{1}{c|}{192}&\textbf{0.163$\pm$1e$^{-3}$}&\textbf{0.250$\pm$1e$^{-3}$}&0.184&0.289&0.199&0.315&0.213&0.316&0.196&0.285&0.201&0.315&0.222&0.334&0.297&0.390\\
				\multicolumn{1}{c|}{}&\multicolumn{1}{c|}{336}&\textbf{0.179$\pm$1e$^{-3}$}&\textbf{0.264$\pm$1e$^{-3}$}&0.198&0.300&0.212&0.329&0.230&0.333&0.209&0.301&0.214&0.329&0.231&0.3383&0.317&0.403\\
				\multicolumn{1}{c|}{}&\multicolumn{1}{c|}{720}&\textbf{0.210$\pm$1e$^{-3}$}&\textbf{0.294$\pm$1e$^{-3}$}&0.220&0.320&0.233&0.345&0.265&0.360&0.245&0.333&0.246&0.355&0.254&0.361&0.338&0.417\\
				\cline{2-18}
				\multicolumn{1}{c|}{}&\multicolumn{1}{c|}{Avg}&\textbf{0.176$\pm$1e$^{-3}$}&\textbf{0.294$\pm$1e$^{-3}$}&0.192&0.295&0.208&0.323&0.229&0.329&0.212&0.300&0.214&0.327&0.227&0.338&0.313&0.401\\
				\hline
				\multicolumn{1}{c|}{\multirow{5}{*}{\rotatebox{90}{Traffic}}}&\multicolumn{1}{c|}{96}&\textbf{0.451$\pm$3e$^{-3}$}&\textbf{0.280$\pm$3e$^{-3}$}&0.593&0.321&0.607&0.392&0.615&0.391&0.650&0.396&0.587&0.366&0.613&0.388&0.798&0.436\\
				\multicolumn{1}{c|}{}&\multicolumn{1}{c|}{192}&\textbf{0.458$\pm$2e$^{-3}$}&\textbf{0.277$\pm$2e$^{-3}$}&0.617&0.336&0.621&0.399&0.601&0.382&0.598&0.370&0.604&0.373&0.616&0.382&0.849&0.481\\
				\multicolumn{1}{c|}{}&\multicolumn{1}{c|}{336}&\textbf{0.477$\pm$2e$^{-3}$}&\textbf{0.278$\pm$2e$^{-3}$}&0.629&0.336&0.622&0.396&0.613&0.386&0.605&0.373&0.621&0.383&0.622&0.337&0.828&0.476\\
				\multicolumn{1}{c|}{}&\multicolumn{1}{c|}{720}&\textbf{0.520$\pm$3e$^{-3}$}&\textbf{0.288$\pm$2e$^{-3}$}&0.640&0.350&0.632&0.396&0.658&0.407&0.645&0.394&0.626&0.382&0.660&0.408&0.854&0.489\\
				\cline{2-18}
				\multicolumn{1}{c|}{}&\multicolumn{1}{c|}{Avg}&\textbf{0.477$\pm$3e$^{-3}$}&\textbf{0.281$\pm$2e$^{-3}$}&0.620&0.336&0.621&0.396&0.622&0.392&0.625&0.383&0.610&0.376&0.628&0.379&0.832&0.471\\
				\hline
				\multicolumn{1}{c|}{\multirow{5}{*}{\rotatebox{90}{Weather}}}&\multicolumn{1}{c|}{96}&\textbf{0.160$\pm$8e$^{-4}$}&\textbf{0.198$\pm$9e$^{-4}$}&0.172&0.220&0.197&0.281&0.182&0.242&0.196&0.255&0.217&0.296&0.266&0.336&0.174&0.252\\
				\multicolumn{1}{c|}{}&\multicolumn{1}{c|}{192}&\textbf{0.203$\pm$9e$^{-4}$}&\textbf{0.239$\pm$1e$^{-3}$}&0.219&0.261&0.237&0.312&0.227&0.287&0.237&0.296&0.276&0.336&0.307&0.367&0.238&0.313\\
				\multicolumn{1}{c|}{}&\multicolumn{1}{c|}{336}&\textbf{0.252$\pm$1e$^{-3}$}&\textbf{0.276$\pm$1e$^{-3}$}&0.280&0.306&0.298&0.353&0.282&0.334&0.283&0.335&0.339&0.380&0.359&0.395&0.287&0.355\\
				\multicolumn{1}{c|}{}&\multicolumn{1}{c|}{720}&\textbf{0.326$\pm$1e$^{-3}$}&\textbf{0.328$\pm$1e$^{-3}$}&0.365&0.359&0.352&0.288&0.352&0.386&0.345&0.381&0.403&0.428&0.419&0.428&0.384&0.415\\
				\cline{2-18}
				\multicolumn{1}{c|}{}&\multicolumn{1}{c|}{Avg}&\textbf{0.235$\pm$1e$^{-3}$}&\textbf{0.260$\pm$1e$^{-3}$}&0.259&0.287&0.271&0.334&0.261&0.312&0.265&0.317&0.309&0.360&0.338&0.382&0.271&0.334\\
				\hline
				\multicolumn{1}{c|}{\multirow{5}{*}{\rotatebox{90}{Exchange}}}&\multicolumn{1}{c|}{96}&0.087$\pm$2e$^{-4}$&0.206$\pm$1e$^{-4}$&0.107&0.234&\textbf{0.085}&\textbf{0.204}&0.116&0.262&0.088&0.218&0.148&0.278&0.197&0.323&0.395&0.474\\
				\multicolumn{1}{c|}{}&\multicolumn{1}{c|}{192}&\textbf{0.171$\pm$2e$^{-4}$}&\textbf{0.295$\pm$1e$^{-4}$}&0.226&0.344&0.182&0.303&0.215&0.359&0.176&0.315&0.271&0.380&0.300&0.369&0.776&0.698\\
				\multicolumn{1}{c|}{}&\multicolumn{1}{c|}{336}&\textbf{0.285$\pm$1e$^{-4}$}&\textbf{0.389$\pm$2e$^{-4}$}&0.367&0.448&0.348&0.428&0.377&0.466&0.313&0.427&0.460&0.500&0.509&0.524&1.029&0.797\\
				\multicolumn{1}{c|}{}&\multicolumn{1}{c|}{720}&\textbf{0.578$\pm$2e$^{-4}$}&\textbf{0.574$\pm$2e$^{-4}$}&0.964&0.746&1.025&0.774&0.831&0.699&0.839&0.695&1.195&0.841&1.447&0.941&2.283&1.222\\
				\cline{2-18}
				\multicolumn{1}{c|}{}&\multicolumn{1}{c|}{Avg}&\textbf{0.280$\pm$2e$^{-4}$}&\textbf{0.366$\pm$2e$^{-4}$}&0.416&0.443&0.410&0.427&0.385&0.447&0.354&0.414&0.519&0.500&0.613&0.539&1.121&0.798\\
				\hline
				\multicolumn{2}{c|}{Improvement}&\multicolumn{2}{c|}{---}&14.8\%&7.7\%&19.1\%&14.3\%&24.3\%&18.5\%&19.4\%&13.8\%&20.8\%&16.0\%&29.2\%&20.8\%&62.3\%&45.6\%\\
				\hline
				\multicolumn{2}{c|}{$1^{st}$ Count}&\multicolumn{2}{c|}{\textbf{56}}&\multicolumn{2}{c|}{3}&\multicolumn{2}{c|}{3}&\multicolumn{2}{c|}{0}& \multicolumn{2}{c|}{0}&\multicolumn{2}{c|}{2}&\multicolumn{2}{c|}{0}&\multicolumn{2}{c}{0}\\
				\hline
	\end{tabular}}}
	\caption{Comparing U-Mixer with SOTA benchmarks on large-scale real-world time-series datasets in long-term forecasting.}
	\label{comparelong}
\end{table*}

\begin{table*}[th]
	\renewcommand{\arraystretch}{1}
	\centering
	\resizebox{\textwidth}{!}{\setlength{\tabcolsep}{0.35mm}{
			\begin{tabular}{c c c c c c c c c c c c c c c c}
				\hline
				\multicolumn{2}{c}{Models}&\textbf{U-Mixer}&TimesNet&N-HiTS&N-BEATS&ETS.&LightTS&FED.&Stationary&Pyra.&In.&LogTrans&Re.&LSTM&TCN\\
				\hline
				\multirow{3}{*}{\rotatebox{90}{Yearly}}&
				\multicolumn{1}{|c|}{SMAPE}&\textbf{13.317$\pm$2e$^{-3}$}&13.387&13.418&13.436&18.009&14.247&13.728&13.717&15.530&14.727&17.107&16.169&176.040&14.920\\
				&\multicolumn{1}{|c|}{MASE}&3.006$\pm$1e$^{-2}$&\textbf{2.996}&3.045&3.043&4.487&3.109&3.048&3.078&3.711&3.418&4.177&3.800&31.033&3.364\\
				&\multicolumn{1}{|c|}{OWA}&\textbf{0.786$\pm$2e$^{-4}$}&\textbf{0.786}&0.793&0.794&1.115&0.827&0.803&0.807&0.942&0.881&1.049&0.973&9.290&0.880\\
				\hline
				\multirow{3}{*}{\rotatebox{90}{Quart.}}&
				\multicolumn{1}{|c|}{SMAPE}&\textbf{9.956$\pm$2e$^{-4}$}&10.100&10.202&10.124&13.376&11.364&10.792&10.958&15.449&11.360&13.207&13.313&172.808&11.122\\
				&\multicolumn{1}{|c|}{MASE}&\textbf{1.156$\pm$3e$^{-3}$}&1.182&1.194&1.169&1.906&1.328&1.283&1.325&2.350&1.401&1.827&1.775&19.753&1.360\\
				&\multicolumn{1}{|c|}{OWA}&\textbf{0.873$\pm$1e$^{-4}$}&0.890&0.899&0.886&1.302&1.000&0.958&0.981&1.558&1.027&1.266&1.252&15.049&1.001\\		
				\hline
				\multirow{3}{*}{\rotatebox{90}{Others}}&
				\multicolumn{1}{|c|}{SMAPE}&\textbf{4.858$\pm$3e$^{-4}$}&4.891&5.061&4.925&7.267&15.880&4.954&6.302&24.786&24.460&23.236&32.491&186.282&7.186\\
				&\multicolumn{1}{|c|}{MASE}&\textbf{3.195$\pm$1e$^{-2}$}&3.302&3.216&3.391&5.240&11.434&3.264&4.064&18.581&20.960&16.288&33.355&119.294&4.677\\
				&\multicolumn{1}{|c|}{OWA}&\textbf{1.015$\pm$1e$^{-4}$}&1.035&1.040&1.053&1.591&3.474&1.036&1.304&5.538&5.879&5.013&8.679&38.411&1.494\\
				\hline
				\multirow{3}{*}{\rotatebox{90}{Avg.}}&\multicolumn{1}{|c|}{SMAPE}&\textbf{11.740$\pm$1e$^{-3}$}&11.829&11.927&11.851&14.718&13.525&12.840&12.780&16.987&14.086&16.018&18.200&160.031&13.961\\
				&\multicolumn{1}{|c|}{MASE}&\textbf{1.575$\pm$1e$^{-2}$}&1.585&1.613&1.599&2.408&2.111&1.701&1.756&3.265&2.718&3.010&4.223&25.788&1.945\\
				&\multicolumn{1}{|c|}{OWA}&\textbf{0.845$\pm$1e$^{-4}$}&0.851&0.861&0.855&1.172&1.051&0.918&0.930&1.480&1.230&1.378&1.775&12.642&1.023\\
				\hline
	\end{tabular}}}
	\caption{Comparing U-Mixer with SOTA benchmarks on M4 datasets in short-term forecasting.}
	\label{compareshort}
\end{table*}
\subsection{Instance Normalization and Learning Objective}
We flatten $\hat{Y_d}$ into $\hat{Y_p}\in \mathbb{R}^{C\times (L+H)}$ by a linear layer. Then we use instance normalization to mitigate the distribution shift effect between the input $X$ and forecasting result. So the output of U-Mixer is $\hat{Y}\in \mathbb{R}^{C\times H}=\hat{Y_p}[:,\tau]\times \sigma_{in}+\mu_{in}$, $\tau=L:L+H$. We choose $L_1$ Loss function to measure the discrepancy between the prediction and the ground truth, and the loss is propagated back from the outputs across the entire model. Compared to the commonly used MSE loss function in time series forecasting tasks, $L_1$ loss function is less sensitive to outliers, which allows the model to achieve more robust performance. The loss of our model is:
\begin{equation}
	\mathcal{L}_{U-Mixer}= \frac{1}{C}\sum_{i=1}^{C}|Y[i,:]-\hat{Y}[i,:]|
\end{equation}
\section{Experiments}
\subsection{Datasets}
We evaluate U-Mixer mainly on six large-scale real-world time-series datasets in long-term forecasting. (1) \textbf{Electricity transformer temperature (ETT)} \cite{Autoformer} data contains the power load features and oil temperature collected from electricity transformers, consisting of seven features. Following the same protocol as Informer \cite{Informer}, we split the data into four datasets: ETTh1, ETTh2, ETTm1, and ETTm2. (2) \textbf{Electricity Consuming Load (ECL)} \cite{Autoformer} data contains the hourly electricity consumption of 321 customers from 2012 to 2014. (3) \textbf{Traffic} \cite{catn} data is a collection of hourly data from California Department of Transportation and describes the occupancy rate of different lanes measured by different sensors on San Francisco highway. (4) \textbf{Exchange} \cite{Exchange} data collects the panel data of daily exchange rates from 8 countries ranging from 1990 to 2016. (5) \textbf{Weather} \cite{PatchTST} data is recorded 21 meteorological indicators collected every 10 minutes from the Weather Station of the Max Planck Biogeochemistry Institute in 2020. (6) \textbf{M4} \cite{TimesNet} is a dataset for the short-term forecasting, which involve 6 subsets: M4-Yearly, M4-Quarterly, M4-Monthly, M4-Weakly, M4-Daily, and M4-Hourly. We divide all datasets into training, validation and testing sets according to the chronological order by the ratio of 6:2:2 for ETT dataset and 7:1:2 for the other datasets.
\subsection{Model Configuration and Metrics}
U-Mixer is implemented through Pytorch and trained on an Nvidia A40 GPU (48GB). The following model configuration is used by default for long-term forecasting: Input sequence length $L=96$, patch length $P=16$, stride $S=8$, forecasting sequence length $H\in \{96, 192, 336, 720\}$, the number of levels $M=3$, and batch size is set to 16. For short-term forecasting the model configuration is: patch length $P=8$, stride $S=4$, and batch size is set to 32. The model training runs 10 epochs, and optimization is performed by Adam. To enhance the reproducibility of the implemented results, we fix random seeds. We employ mean square error (MSE) and mean absolute error (MAE) for long-term forecasting evaluation. Following the N-BEATS \cite{N-BEATS}, we also employ the symmetric mean absolute percentage error (SMAPE), mean absolute scaled error (MASE) and overall weighted average (OWA) for short-term forecasting.
\subsection{SOTA Benchmarks}
We compare U-Mixer with the following 16 SOTA methods from five different categories: (1) \textbf{RNN-based models}: LSTM \cite{lstm} and LSSL \cite{LSSL}. (2) \textbf{MLP-based models}: LightTS \cite{LightTS} and DLinear \cite{dlinear}. (3) \textbf{CNN-based models}: TCN \cite{TCN} and TimesNet \cite{TimesNet}. (4) \textbf{Transformer-based models}: LogTrans \cite{LogTrans}, Reformer \cite{Reformer}, Informer \cite{Informer}, Pyraformer \cite{Pyraformer}, Autoformer \cite{Autoformer}, FEDformer \cite{Fedformer}, Non-stationary Transformer \cite{Non-stationary}, and ETSformer \cite{Etsformer}. (5) \textbf{Decomposition-based models}: N-BEATS \cite{N-BEATS} and N-HiTS \cite{NHITS}.
\subsection{Overall Comparison}
Table \ref{comparelong} presents the long-term forecasting performance measured by MSE and MAE, and Table \ref{compareshort} shows the short-term forecasting performance measured by SMAPE, MASE, and OWA. We can observe that U-Mixer balances short- and long-term forecasing well and achieves the best short- and long-term forecasting performance. Out of the 64 long-term forecastings made on 8 datasets, we achieve the best results in 56 cases and brings 14.5\%/7.7\% improvements on MSE/MAE to the existing best results. For short-term forecasting on the M4 dataset, we achieve nearly all optimal outcomes. Specially, we found that RNN-based methods exhibites significantly poorer performance. For long-term forecasting, there is not a substantial difference among MLP-based, CNN-based, and Transformer-based methods. Apart from our model, the Decomposition-based methods demonstrates a leading position in short-term prediction. 
\begin{table}[t]
	\renewcommand{\arraystretch}{0.94}
	\centering
	\begin{tabular}{c|c|c|c|c}
		\hline
		\multicolumn{2}{c|}{Models}&\textbf{U-Mixer}&w/o UE&w/o SC\\
		\hline
		\multirow{2}{*}{ETTh2}&\multicolumn{1}{c|}{MSE}&\textbf{0.381}&0.405&0.400\\
		&\multicolumn{1}{c|}{MAE}&\textbf{0.399}&0.427&0.421\\
		\hline
		\multirow{2}{*}{Weather}&\multicolumn{1}{c|}{MSE}&\textbf{0.235}&0.252&0.254\\
		&\multicolumn{1}{c|}{MAE}&\textbf{0.260}&0.285&0.287\\
		\hline
		\multirow{3}{*}{M4}&\multicolumn{1}{c|}{SMAPE}&\textbf{11.740}&11.817&11.808\\
		&\multicolumn{1}{c|}{MASE}&\textbf{1.575}&1.582&1.587\\
		&\multicolumn{1}{c|}{OWA}&\textbf{0.845}&0.857&0.853\\
		\hline
	\end{tabular}
	\caption{Results of ablation study on ETTh2, Weather and M4 datasets.}
	\label{ablation}
\end{table}
\begin{figure}[h]
	\centering
	\includegraphics[width=0.5\textwidth]{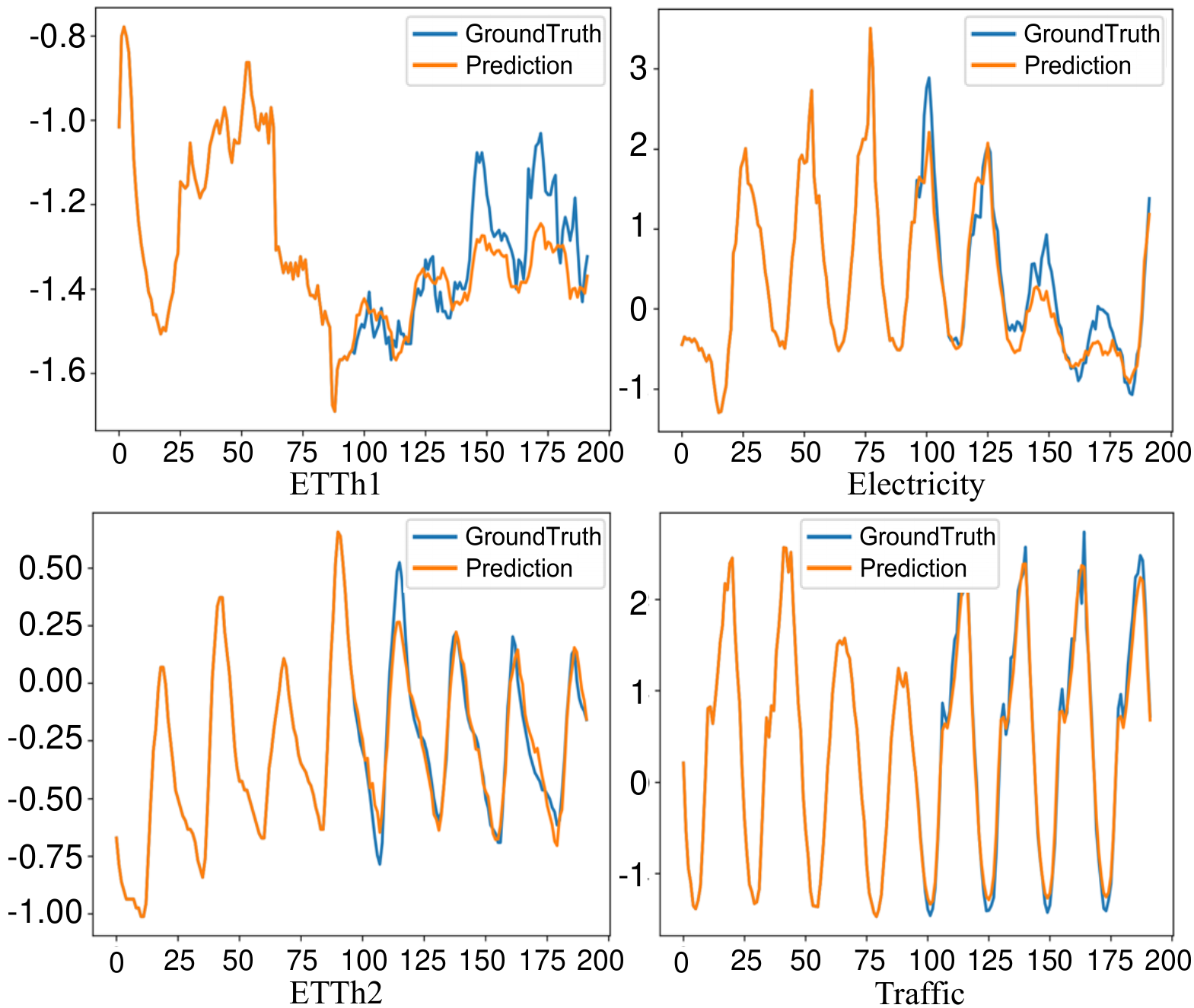} 
	\caption{Visualization of forecasting results on multiple datasets by U-Mixer.}
	\label{showcases}
\end{figure}
\begin{figure}[h]
	\centering
	\includegraphics[width=0.47\textwidth]{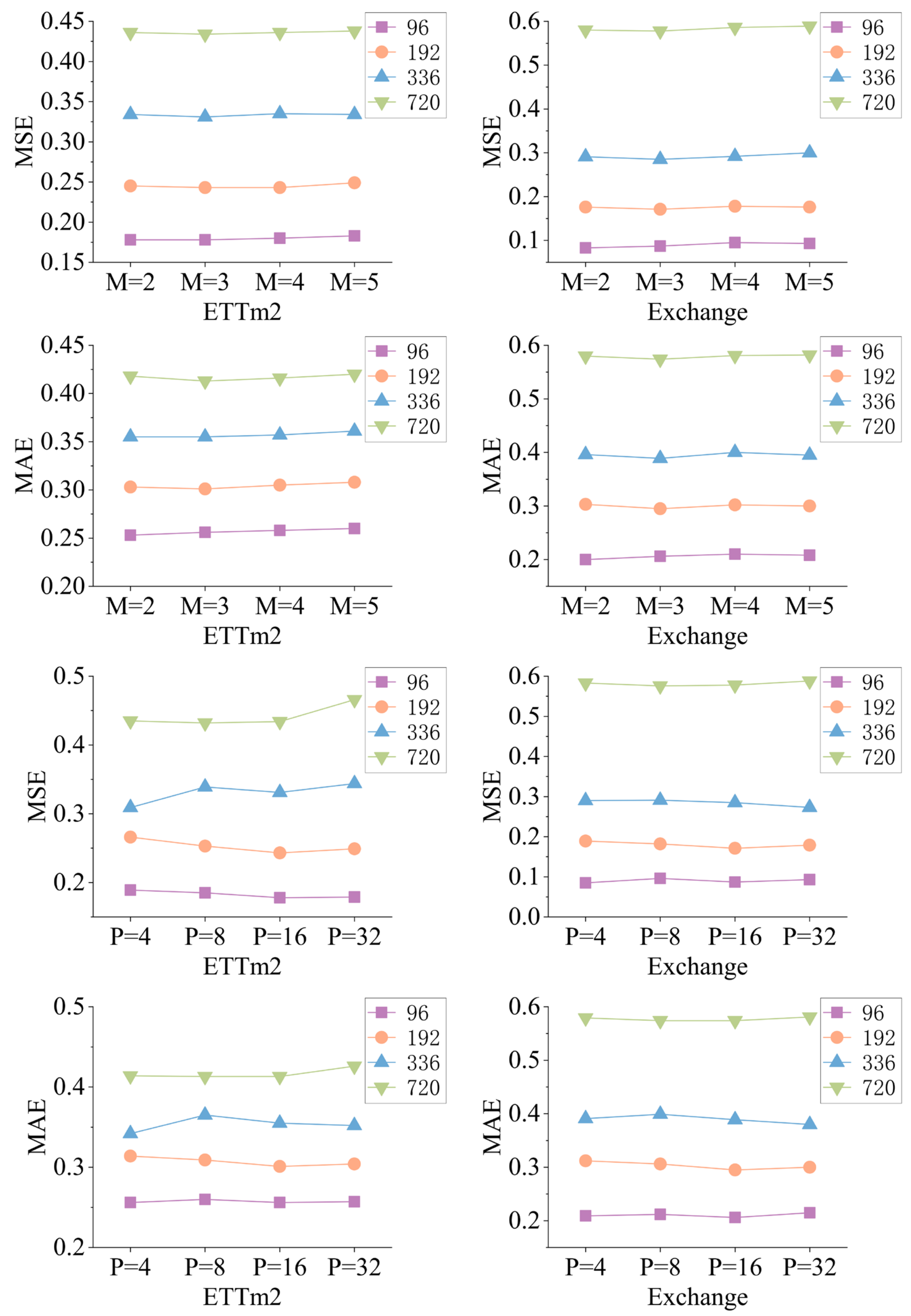} 
	\caption{Performance comparison on varying.}
	\label{para}
\end{figure}
\subsection{Ablation Study}
To better evaluate the Unet encoder-decoder (UE) and stationarity correction (SC), we conduct supplementary experiments with ablation consideration. Here\textbf{ w/o UE} and \textbf{w/o SC} are variants of U-Mixer. In w/o UE, we set the Unet encoder-decoder level $M=0$. In w/o SC, we remove the stationarity correction process. We set the batch size as 16, input sequence length as 96, horizon in $\{96, 192, 336, 720\}$ and use the same parameter setting in all ablation experiments. The average results are shown in Table \ref{ablation}. All metrics of w/o UE and w/o SC are significantly worsed than U-Mixer. The complete U-Mixer can obtain the best results, demonstrating removing any components of it will affect the effect. The Unet encoder-decoder and stationarity correction are necessary and effective. 
\subsection{Showcases}
To further show the forecasting performance of U-Mixer, we visualize the results on ETTh1, Electricity, ETTh2, and Traffic datasets. As shown in Figure \ref{showcases}, the ETTh2 and Traffic datasets exhibit more clear periodicity patterns, which our method can efficiently forecasting the ground truth. This demonstrates the strong capability of U-Mixer in capturing period. In contrast, the patterns of ETTh1 and Electricity are relatively less obvious. Our method manages to capture their periodicity to a reasonable extent, and also makes certain forecastings on their trend.  It is verified the robustness of  U-Mixer performance among various data characteristics.
\subsection{Parameter Sensitivity}
We choose the perform the sensitivity analysis of U-Mixer on ETTm2 and Exchange datasets by varying the number of level $M$ and the patch length $P$. As shown in Figure \ref{para}, it can be concluded that our model is not highly sensitive to parameter $M$. Overall, the optimal results are achieved when $M$ is set to 3. Additionally, with an increase in $M$, the training time of the model also increases. Therefore, we choose $M=3$ as the parameter for our model. For the parameter $P$, we can tell our model exhibits a higher sensitivity to parameter $P$ on the ETTm2 dataset, while the sensitivity is lower on the Exchange dataset. The value of $P$ also influences the training time. When $P$ is set to 8 or 16, the training time is relatively shorter, with the outcome of $P=16$ surpassing that of $P=8$. Consequently, we set $P=16$ as the designated value.
\section{Conclusion}
In this paper, we study the non-stationarity challenge in time series forecasting task and propose a novel forecasting model U-Mixer. We combine the Unet and Mixer architecture to capture the local dependencies between patches and channels separately of different levels to obtain the comprehensive representations of time series. More importantly, we propose a stationarity correction method to handle the distribution shift caused by non-stationarity while preserving the temporal dependencies. Our model demonstrates state-of-the-art short- and long-term forecasting performance on six real-world datasets, and the overall superiority of various experiments further verifies the effectiveness and robustness of it.

\section{Acknowledgements}
Supported by the National Natural Science Foundation of China (NSFC) Joint Fund with Zhejiang Integration of Informatization and Industrialization under Key Project (Grant No.U22A2033) and NSFC (Grant No.62072281).

\bibliography{UMixer}

\end{document}